\pdfoutput=1

\documentclass[11pt]{article}

\usepackage[final]{acl}

\usepackage{times}
\usepackage{latexsym}

\usepackage[T1]{fontenc}

\usepackage[utf8]{inputenc}

\usepackage{microtype}

\usepackage{inconsolata}

\usepackage{graphicx}
\usepackage{multirow}
\usepackage{enumitem}
\usepackage{booktabs}
\usepackage{listings}
\usepackage{amsmath, bm}
\usepackage{pifont}
\usepackage{hyperref}
\usepackage{adjustbox}
\usepackage{subcaption}
\usepackage{tabularray}
\usepackage{relsize}

\usepackage[colorinlistoftodos]{todonotes}
\title{Event-based evaluation of abstractive news summarization}

\author{Huiling You$^1$, Samia Touileb$^2$, Erik Velldal$^1$, \and Lilja Øvrelid$^1$ \\
         $^1$University of Oslo\\
         $^2$University of Bergen \\ 
         {\tt \{huiliny, erikve, liljao\}@ifi.uio.no} \\
         {\tt samia.touileb@uib.no}
}

\begin{document}
\maketitle
\begin{abstract}
An abstractive summary of a news article contains its most important information in a condensed version. The evaluation of automatically generated summaries by generative language models relies heavily on human-authored summaries as gold references, by calculating overlapping units or similarity scores. News articles report events, and ideally so should the summaries. In this work, we propose to evaluate the quality of abstractive summaries by calculating overlapping events between generated summaries, reference summaries, and the original news articles. We experiment on a richly annotated Norwegian dataset comprising both events annotations and summaries authored by expert human annotators. Our approach provides more insight into the event information contained in the summaries.
\end{abstract}

\section{Introduction}

A summary of a news article provides a condensed version of its main content \citep{el2021automatic}. One of the primary practical applications of large language models (LLMs) is generating concise text summaries, and many news publishers in Norway have already integrated LLM-generated summaries into their articles. However, assessing the quality and accuracy of these summaries remains a challenge. Current evaluation metrics compare generated summaries to ideal summaries created by humans, in terms of overlapping words/units, such as ROUGE-L \citep{lin-2004-rouge}, or semantic similarity, such as BERTScore \citep{Zhang*2020BERTScore:}. However, these metrics provide limited information on the semantic content of the summaries themselves.

With the increasing usage of LLMs for text generation, there has been a growing number of studies on evaluating the factuality of these texts from the perspective of contained information, such as FACTSCORE \citep{min-etal-2023-factscore}. For summarization, \citet{zhang-bansal-2021-finding} propose to use semantic triplet units as a judgment of the semantic content units in generated texts, and \citet{liu-etal-2023-revisiting} also propose a similar protocol based on semantic units, named Atomic Content Units. Inspired by event extraction (EE), a NLP task that extracts event information from unstructured texts into structured forms \citep{doddington-etal-2004-automatic}, we propose to analyze the quality of news article summaries by comparing the overlapping events between generated summaries, reference summaries, and the source articles. By using structured event information, we provide more insight into both the generated summaries and human-authored summaries. We experiment on a Norwegian dataset with rich annotations both for events (EDEN \citep{touileb-etal-2024-eden}), and summaries (NorSumm \citep{touileb2025norsumm}), and demonstrate the usefulness of the proposed event-based evaluation metric which is grounded in the overlap of identified events.

\section{Event-overlap}

Our proposed metric calculates the degree of overlapping events between summaries (generated and human-authored) and the source texts. First, an event extraction system is used to extract events from summaries and source articles. Second, standard event extraction evaluation metrics are adapted and applied to calculate the actual event overlaps.

\subsection{Event extraction}

\begin{figure}
    \centering
    \includegraphics[width=0.98\linewidth]{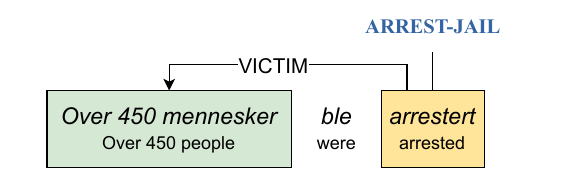}
    \caption{Example of a sentence with event annotation. The \texttt{ARREST-JAIL} event has the trigger ``arrested'', and the \texttt{VICTIM} argument is ``Over 450 people''.}
    \label{fig:event-eg}
\end{figure}

An event \citep{doddington-etal-2004-automatic} contains four key elements: 1) \textbf{event type} is the specific type of event defined within an ontology; 2) \textbf{event trigger} is the word(s) in the text that describes the event; 3) \textbf{event argument} is the attribute and actual participant of an event in the text; 4) \textbf{argument role} is the role played by an argument in the specific event. Figure~\ref{fig:event-eg} shows an example of a Norwegian sentence annotated for an \texttt{ARREST-JAIL} event with ``arrested'' as the event trigger, and a \texttt{VICTIM} argument ``Over 450 people''. We use an existing event extraction system NorEventGen \citep{noreventgen} to obtain event information in these structured formats.

We perform event extraction on three different texts: 1) model-generated summaries; 2) human-authored summaries; and 3) original news articles.

\subsection{Event-overlap analysis}

Our event-overlap metric is adapted from the classical evaluation metrics of event extraction \citep{lin-etal-2020-joint, nguyen-etal-2021-cross}, as follows: an event trigger is correctly identified (Trg-I) if its offsets match a reference trigger, and correctly classified (Trg-C) if its event type also matches a reference trigger; An argument is correctly identified (Arg-I) if its offsets match a reference argument, and correctly classified (Arg-C) if its argument role also matches the reference argument. 

Since an abstractive summary does not perform text extraction from the source article, we do not expect a perfect match between an event trigger / argument from the summary and one from the article. As an alternative, we use BERTScore \citep{Zhang*2020BERTScore:} as a reference to check if two pieces of texts are similar.\footnote{We use a heuristic threshold of 0.7. If the BERTScore is larger than 0.7, two text snippets will be considered similar, the same as perfect match in event extraction metric.} Unlike in event extraction, we prioritize the labels, namely event type and argument role. We do not take trigger word(s) into account, because the event type information itself is sufficient, and unlike event arguments, which are named entities, trigger words are more often rephrased with a different choice of words in summaries. With the corresponding adaptation, our proposed event-overlap metric calculates the following three categories of scores:
\begin{itemize}[noitemsep]
    \item An event type (eType-C) overlaps if it exists in both lists of extracted events.
    \item An argument role (Role-C) overlaps if the event type and argument role overlap.
    \item An argument (Arg-C) overlaps if the event type, argument role, and argument word(s) overlap.
\end{itemize}

The Precision (P), Recall (R), and F1 scores of each category are calculated. The final event-overlap score is an aggregated score of the three categories of scores: $\rm \textbf{Event\textrm{-}overlap}=\textbf{Average}([eType\textrm{-}C, Role\textrm{-}C, Arg\textrm{-}C])$. Depending on the event overlap of different texts, different scores are used:
\begin{itemize}[noitemsep]
    \item \textbf{Event-overlap between summaries}: the final event-overlap score is the average Recall scores of eType-C, Role-C, and Arg-C. Recall scores prioritize the events that are in the gold summaries.
    \item \textbf{Event-overlap between summaries and original articles}: the final event-overlap score is the average Precision scores of eType-C, Role-C, and Arg-C. Precision scores provide evaluation of identified events in the summaries that are also present in the original articles.
\end{itemize}

\section{Experimental setup}

\paragraph{Datasets} We use two recently released datasets: the Norwegian event detection dataset EDEN \citep{touileb-etal-2024-eden} and the human-authored summaries of Norwegian news articles dataset NorSumm \citep{touileb2025norsumm}. The source articles of NorSumm are a subset of EDEN. These parallel annotations of events and summaries make it possible to evaluate our approach and contrast gold vs predicted event information on gold vs generated summaries. More concretely, we use the test set of NorSumm, which contains 33 news articles, each coupled with three unique human-authored summaries.

\paragraph{LLMs} 
For automatic summarization, we evaluate a range of Norwegian and Nordic open-source pretrained and instruction-finetuned decoder-only LLMs: Llama-3-8B-instruct,\footnote{\url{https://huggingface.co/AI-Sweden-Models/Llama-3-8B-instruct}} Llama-3-8B,\footnote{\url{https://huggingface.co/AI-Sweden-Models/Llama-3-8B}} 
Meta-Llama-3-8B-Instruct\footnote{\url{https://huggingface.co/meta-llama/Meta-Llama-3-8B-Instruct}}, Mistral-Nemo-Instruct-2407,\footnote{\url{https://huggingface.co/mistralai/Mistral-Nemo-Instruct-2407}} Normistral-11b-warm\footnote{\url{https://huggingface.co/norallm/normistral-11b-warm}}, and Normistral-7b-warm-instruct.\footnote{\url{https://huggingface.co/norallm/normistral-7b-warm-instruct}} All the LLMs are available via HuggingFace.\footnote{\url{https://huggingface.co/models}} We use the same prompts as in the NorSumm evaluation \citep{touileb2025norsumm} to generate summaries, and keep only one summary that has highest average score of ROUGE-L and BERTScore values for each model.

\paragraph{Event extraction system} We use a generative event extraction system NorEventGen \citep{noreventgen} to identify and extract events from both the original articles and the summaries. NorEventGen is trained on EDEN, and holds the current SOTA results. The system performs sentence-level extraction. In our experiments, both the original articles and the summaries are first split into sentences, and then event prediction is performed on each of the sentences.

\section{Results and discussion}

{\footnotesize
\begin{table*}[h!]
\centering
\begin{adjustbox}{max width=0.95\textwidth}
\begin{tabular}{l|ccc|ccc|ccc|c}
\toprule
\multirow{2}{*}{\textbf{Summary}} & \multicolumn{3}{c|}{\textbf{eType-C}}  & \multicolumn{3}{c|}{\textbf{Role-C}} & \multicolumn{3}{c|}{\textbf{Arg-C}} & \multirow{2}{*}{\textbf{Event-overlap}} \\

&\textbf{P} & \textbf{R} & \textbf{F1} & \textbf{P} & \textbf{R} & \textbf{F1} & \textbf{P} & \textbf{R} & \textbf{F1} &  \\ \midrule

Human-authored & 90.7 & 13.4 & 23.4 & 84.7 & 13.2 & 22.8 & 68.2 & 10.7 & 18.4 & 81.2 \\ \midrule

Llama-3-8B-instruct & 93.3 & 8.3 & 15.3 & 87.3 & 6.8 & 12.6	& 70.4 & 5.5 & 10.2 & 83.7 $_{(6)}$ \\

Llama-3-8B & \textbf{98.4} & 12.1 & 21.5 & 89.2 & 10.8 & 19.3 & 81.1	& 9.9 & 17.6 & 89.6 $_{(2)}$ \\



Meta-Llama-3-8B-Instruct & 97.8 & 8.9 & 16.3 & 90.0 & 7.9 & 14.5 & 76.3 & 6.7 & 12.3 & 88.0 $_{(3)}$ \\ 

Mistral-Nemo-Instruct-2407 & 98.0 & 9.5 & 17.3 & 87.1 & 8.1 & 14.8 & 69.4 & 6.5 & 11.8 & 84.8 $_{(4)}$ \\

Normistral-11b-warm & 96.7 & \textbf{17.2} & \textbf{29.2} & \textbf{90.8} & 16.2 & 27.5 & \textbf{82.2} & \textbf{14.7} & \textbf{24.9} & \textbf{89.9} $_{(1)}$ \\

Normistral-7b-warm-instruct & 94.6 & \textbf{17.2} & 29.1 & 88.5 & \textbf{16.9} & \textbf{28.3} & 69.5 & 13.3 & 22.3 & 84.2 $_{(5)}$\\

\bottomrule
         
    \end{tabular}
    \end{adjustbox}
    \caption{Event-overlap between summaries and the original articles, with event prediction is performed with NorEventGen. The subscripts indicate the corresponding ranking of the model based on the score.}
    \label{tab:gen_vs_pred}
\end{table*}}

{\footnotesize
\begin{table*}[h!]
\centering
\begin{adjustbox}{max width=0.95\textwidth}
\begin{tabular}{l|ccc|ccc|ccc|c}
\toprule
\multirow{2}{*}{\textbf{Summary}} & \multicolumn{3}{c|}{\textbf{eType-C}}  & \multicolumn{3}{c|}{\textbf{Role-C}} & \multicolumn{3}{c|}{\textbf{Arg-C}} & \multirow{2}{*}{\textbf{Event-overlap}} \\

&\textbf{P} & \textbf{R} & \textbf{F1} & \textbf{P} & \textbf{R} & \textbf{F1} & \textbf{P} & \textbf{R} & \textbf{F1} & \\ \midrule

Human-authored & 74.2 & 13.1 & 22.4 & 69.4 & 11.9 & 20.4 & 59.2	& 10.2 & 17.4 & 67.6 \\ \midrule

Llama-3-8B-instruct & 84.4 & 9.0 & 16.2 & 76.1 & 6.5 & 12.0 & 66.2 & 5.7 & 10.5 & 75.6 $_{(4)}$\\

Llama-3-8B & 82.3 & 12.1 & 21.0 & 76.6 & 10.3 & 18.1 & 68.5 & 9.2 & 16.2 & 75.8 $_{(3)}$\\



Meta-Llama-3-8B-Instruct & \textbf{87.0} & 9.5 & 17.1  & 82.5 & 8.0 & 14.6 & \textbf{75.0} & 7.3 & 13.3 & \textbf{81.5} $_{(1)}$\\ 

Mistral-Nemo-Instruct-2407 & 83.7 & 9.7 & 17.4 & \textbf{83.5} & 8.6 & 15.6 & 74.1 & 7.6	 & 13.8 & 80.4 $_{(2)}$\\

Normistral-11b-warm & 80.0 & 17.0 & 28.1 & 77.3 & 15.3	& 25.5 & 69.3 & \textbf{13.7} & \textbf{22.9} & 75.5 $_{(5)}$\\

Normistral-7b-warm-instruct & 87.0	& \textbf{18.9} & \textbf{31.1} & 77.0 & \textbf{16.2} & \textbf{26.8} & 59.8 & 12.6 & 20.8 & 74.6 $_{(6)}$\\

\bottomrule
         
    \end{tabular}
    \end{adjustbox}
    \caption{Event-overlap between summaries (predicted events) and the original articles (gold events). The subscripts indicate the corresponding ranking of the model based on the score.}
    \label{tab:gen_vs_gold}
\end{table*}}

We here present the analysis of our event-overlap metric on the test set of NorSumm. We first present the event-overlap between summaries and the original articles; we then present the event-overlap between generated summaries and human-authored summaries. Finally, we discuss the overall picture summarizing event-overlap scores.

\subsection{Event-overlap between summaries and the original articles}

Table \ref{tab:gen_vs_pred} shows the event-overlap between the summaries (both human-authored and generated) and the original articles. As the results show, both generated summaries and human-authored summaries generally discuss events that are described in the original articles, and there are always fewer events in the summaries. As the Precision scores of eType-C are always above 90\%, it is rare for events that are not discussed in the source article to be mentioned in the summary, which is especially true for generated summaries. The Recall scores of eType-C are much lower, meaning that there are far fewer events in the summaries; the number of events varies considerably among generated summaries. The Precision scores of Role-C and Arg-C show that events are discussed with different levels of detail in the summaries compared to the news articles. Similarly, the event-overlap metric shows that Normistral-11b-warm is the best-performing model, but the summaries generated by Llama-3-8B and Normistral-7b-warm-instruct also produce relatively good results with each of the fine-grained metrics.

Table~\ref{tab:summ_event_stats} provides detailed event statistics of both human-authored and generated summaries, together with event information of the original articles. In general, there are always fewer events in the summaries as compared to in the original articles, which is expected. Human annotators have rather high agreement on event numbers, but the number of argument roles vary quite a lot, meaning they tend to describe the events with varied details when writing the summaries. For model-generated summaries, some describe considerably more events than others; the summaries generated by Normistral-7b-warm-instruct contain twice the number of events compared with the summaries generated by Llama-3-8B-instruct.

{\footnotesize
\begin{table*}[h!]
    \centering
    \begin{adjustbox}{max width=0.9\textwidth}
    \begin{tabular}{l|cccc}
    \toprule
    \textbf{Summary} & \textbf{\#Events} & \textbf{\#Roles} & \textbf{\#Event types} & \textbf{\#Role types} \\ \midrule
    Annotator$_1$ & 77 & 156 & 17 & 23 \\
    Annotator$_2$ & 77 & 146 & 16 & 20 \\
    Annotator$_3$ & 71 & 126 & 16 & 24 \\ \midrule
    Llama-3-8B-instruct & 45 & 71 & 13 & 17 \\
    Llama-3-8B & 62 & 111 & 14 & 19 \\
    Meta-Llama-3-8B-Instruct & 46 & 80 & 14 & 20 \\
    Mistral-Nemo-Instruct-2407 & 49 & 85 & 12 & 19 \\
    Normistral-11b-warm & 90 & 163 & 15 & 20 \\
    Normistral-7b-warm-instruct & 92 & 174 & 15 & 23 \\ \midrule
    Gold events in original articles & 423 & 826 & 23 & 25 \\
    Predicted events in original articles & 506 & 918 & 23 & 25 \\ 
    \bottomrule
    \end{tabular}
    \end{adjustbox}
    \caption{Event statistics of human-authored summaries by three different annotators and generated summaries by different models. Events are predicted with the selected event extraction system.}
    \label{tab:summ_event_stats}
\end{table*}}

Instead of predicted events, we can also assess the influence of event detection accuracy and compare the gold event annotation of the original articles to calculate the event-overlap scores. As Table~\ref{tab:gen_vs_gold} shows, the event-overlap scores are still relatively high, similar to using predicted events of the articles. The drops in scores are expected, because the event extraction model is not perfect and less frequent events are annotated, which would normally not be included in the summary.

With gold events, the ranking of the models turns out to be different from when predicted events are used; summaries generated by Meta-Llama-3-8B-Instruct have the highest event-overlap score with the original articles, instead of Normistral-11b-warm. However, the top-performing models remain quite similar.


\subsection{Event-overlap between summaries}

{\footnotesize
\begin{table*}[h!]
\centering
\begin{adjustbox}{max width=0.95\textwidth}
\begin{tabular}{lll|ccc|ccc|ccc|c}
\toprule
\multirow{2}{*}{\textbf{Model}} & \multirow{2}{*}{\textbf{ROUGE-L}} & \multirow{2}{*}{\textbf{BERTScore}} & \multicolumn{3}{c|}{\textbf{eType-C}} & \multicolumn{3}{c|}{\textbf{Role-C}} & \multicolumn{3}{c|}{\textbf{Arg-C}} & \multirow{2}{*}{\textbf{Event-overlap}} \\

& & & \textbf{P} & \textbf{R} & \textbf{F1} & \textbf{P} & \textbf{R} & \textbf{F1} & \textbf{P} & \textbf{R} & \textbf{F1}  & \\ \midrule

Llama-3-8B-instruct & 24.5 $_{(6)}$ & 72.1 $_{(6)}$ & 74.1 & 44.6 & 55.7 & 
 58.2 & 29.4 & 39.0 & 45.1 & 22.9 & 30.3 & 32.3 $_{(6)}$\\

Llama-3-8B & 36.7 $_{(3)}$ & 73.3 $_{(4)}$ & 74.7 & 61.9 & 67.7 & 61.3 & 48.0 & 53.7 & 44.7 & 35.0 & 39.2 & 48.3 $_{(3)}$\\



Meta-Llama-3-8B-Instruct & 28.8 $_{(5)}$ & 75.2 $_{(2)}$ & \textbf{75.4} & 46.3 & 57.3 & \textbf{62.5} & 35.3 & 45.0 & \textbf{52.9} & 29.8 & 38.1 & 37.1 $_{(4)}$\\ 

Mistral-Nemo-Instruct-2407 & \textbf{41.1} $_{(1)}$ & \textbf{75.8} $_{(1)}$ & 67.4 & 43.9 & 53.2 & 55.7 & 33.0 & 41.4 & 45.5 & 27.0 & 33.8 & 34.6 $_{(5)}$\\

Normistral-11b-warm & 34.9 $_{(4)}$ & 73.1 $_{(5)}$ & 70.4 & \textbf{84.6} & \textbf{76.8} & 55.6 & \textbf{63.9} & \textbf{59.4} & 40.3 & 46.1 & \textbf{42.9} & \textbf{64.9} $_{(1)}$\\

Normistral-7b-warm-instruct & 37.8 $_{(2)}$& 73.7 $_{(3)}$ & 64.5 & 79.2 & 71.1 & 51.0 & 62.6 & 56.1 & 37.9 & \textbf{46.5} & 41.7 & 62.8 $_{(2)}$\\

\bottomrule
         
    \end{tabular}
    \end{adjustbox}
    \caption{Event-overlap between generated summaries and human-authored summaries. The subscripts indicate the corresponding ranking of the model based on the score.}
    \label{tab:gen_vs_summ}
\end{table*}}

{\footnotesize
\begin{table*}[h!]
\centering
\begin{adjustbox}{max width=0.95\textwidth}
\begin{tabular}{l|l}
\toprule
\multirow{4}{*}{\textbf{Human-authored}} & \underline{Tommy Sharif} sikret seg \underline{“Diamanten”, toppen av det historiske} \\
& \underline{Holmenkollen-tårnet}, før nettauksjonen ble avsluttet \underline{kl 16.30 på søndag}. \\
& \textit{Tommy Sharif secured the “Diamond”, the top of the historic} \\
& \textit{Holmenkollen Tower, before the online auction ended at 4:30 p.m. on Sunday.} \\ \hline

\multirow{4}{*}{\textbf{Generated}} &  \underline{Tommy Sharif} sikret seg vinnerbudet på \underline{«Diamanten»} \\
& \underline{på Holmenkollen-tårnet} da nettauksjonen ble avsluttet \underline{søndag}. \\
& \textit{Tommy Sharif secured the winning bid for the “Diamond”} \\
& \textit{on the Holmenkollen Tower when the online auction ended on Sunday.} \\

\bottomrule
         
    \end{tabular}
    \end{adjustbox}
    \caption{Example sentence describing the same event, taken from a human-authored summary and a summary generated by Normistral-11b-warm.}
    \label{tab:summ_example}
\end{table*}}

\begin{table*}[h!]
    \centering
    \begin{adjustbox}{max width=0.95\textwidth}
    \begin{tabular}{l|l}
    \toprule
    \multirow{3}{*}{\textbf{Human-authored}}  &  ARREST-JAIL, ATTACK, BE-BORN, CONVICT, DEMONSTRATE, DIE, ELECT, END-ORG \\ 
    & END-POSITION, INJURE, MEET, PHONE-WRITE, START-ORG, START-POSITION \\
    & TRANSFER-MONEY, TRANSFER-OWNERSHIP, TRANSPORT, TRIAL-HEARING \\
    \midrule
    \multirow{3}{*}{\textbf{Generated}} &  ARREST-JAIL, ATTACK, BE-BORN, CHARGE-INDICT, CONVICT, DEMONSTRATE, DIE, ELECT \\
    & END-ORG, END-POSITION, EXECUTE, FINE, INJURE, MEET, PHONE-WRITE, START-ORG \\
    & START-POSITION, TRANSFER-MONEY, TRANSFER-OWNERSHIP, TRANSPORT, TRIAL-HEARING \\
    \bottomrule
    \end{tabular}
    \end{adjustbox}
    \caption{Event types in human-authored summaries and generated summaries.}
    \label{tab:etype}
\end{table*}

Table~\ref{tab:gen_vs_summ} shows the event-overlap between model-generated summaries and human-authored summaries. As the event-overlap scores show, the proportion of shared events in generated summaries with reference summaries varies across the various models. In general, eType-C scores are much higher than Role-C and Arg-C scores, indicating that the same events are discussed with different details. Table~\ref{tab:summ_example} presents an example of a \texttt{TRANSFER-OWNERSHIP} event described in a human-authored summary and a model-generated summary; the human annotator provides more detail about the \texttt{ARTIFACT}, of which the ownership is transferred, and the \texttt{TIME} of the event, but the model stresses that the \texttt{BUYER} gets a winning bid in the auction.

In terms of event types, there are much fewer event types in the summaries. The event ontology of EDEN defines 34 event types, but only half of the event types exist in the reference summaries and even fewer in generated summaries. As such, only certain event types are often considered as main event types, which are then described in the summary. Table~\ref{tab:etype} lists all the event types that are described in all human-authored summaries and generated summaries, corresponding to 21 and 18 event types.

Compared to ROUGE-L and BERTScore, the standard summarization evaluation metrics, our event-overlap scores result in slightly different rankings of model performances. According to ROUGE-L and BERTScore, the best-performing model is  Mistral-Nemo-Instruct-2407, but our event-overlap metric would identify Normistral-11b-warm as the best-performing model.

\subsection{Event-overlap: a combined picture}

By analyzing the event-overlap scores between model-generated summaries and their corresponding human-authored counterparts, alongside the event-overlap scores between both types of summaries and the original articles, we can gain deeper insight into how each summarization approach captures the core content of the articles. These event-overlap scores, as presented in Table \ref{tab:gen_vs_summ} and \ref{tab:gen_vs_pred}, reveal a notable trend: summaries generated by LLMs often focus on different events within the article compared to those emphasized by human writers. This pattern holds consistently across all the LLMs evaluated in the study. LLMs and human summarizers tend to have different judgments on what constitutes the main events or key points in a news article, showing that LLMs struggle to accurately identify and convey the main story in complex, real-world texts like news articles.

\section{Conclusion}

In this article, we introduce a new approach for evaluating abstractive summaries using event identification information. Our proposed event-overlap metric quantifies shared events between generated summaries, human-authored summaries, and the original news articles, offering more insight into the event information of the summaries. In conjunction with standard summarization evaluation metrics, our event-overlap metric adds a valuable dimension to assessing the quality of LLM generated summaries. Experiments conducted on NorSumm, a richly annotated Norwegian dataset, demonstrate the effectiveness and practicality of our method. Our approach is also easily adaptable to other datasets and languages.

\section*{Limitations}

Our work has the following limitations: 1) we only experiment on a small Norwegian dataset, and the event annotation is on a sentence level, but a summary is a condensed version of the entire article; 2) the selected set of generative LLMs is limited; 3) we make a considerable change to the perfect match of argument words in the original event extraction evaluation metric, and our new equivalent using BERTScore with a heuristic value of 0.7 as threshold, needs further experiments; 4) our event-overlap metric is limited by the event extraction system used, and current event extraction systems are far from being perfect.

\section*{Acknowledgments}
 This work was supported by industry partners and the Research Council of Norway with funding to MediaFutures: Research Centre for Responsible Media Technology and Innovation, through the centers for Research-based Innovation scheme, project number 309339.


\bibliography{custom,anthology}

\begin{thebibliography}{12}
\providecommand{\natexlab}[1]{#1}

\bibitem[{Doddington et~al.(2004)Doddington, Mitchell, Przybocki, Ramshaw, Strassel, and Weischedel}]{doddington-etal-2004-automatic}
George Doddington, Alexis Mitchell, Mark Przybocki, Lance Ramshaw, Stephanie Strassel, and Ralph Weischedel. 2004.
\newblock \href {https://aclanthology.org/L04-1011/} {The automatic content extraction ({ACE}) program {--} tasks, data, and evaluation}.
\newblock In \emph{Proceedings of the Fourth International Conference on Language Resources and Evaluation ({LREC}`04)}, Lisbon, Portugal. European Language Resources Association (ELRA).

\bibitem[{El-Kassas et~al.(2021)El-Kassas, Salama, Rafea, and Mohamed}]{el2021automatic}
Wafaa~S El-Kassas, Cherif~R Salama, Ahmed~A Rafea, and Hoda~K Mohamed. 2021.
\newblock Automatic text summarization: A comprehensive survey.
\newblock \emph{Expert systems with applications}, 165:113679.

\bibitem[{Lin(2004)}]{lin-2004-rouge}
Chin-Yew Lin. 2004.
\newblock \href {https://aclanthology.org/W04-1013/} {{ROUGE}: A package for automatic evaluation of summaries}.
\newblock In \emph{Text Summarization Branches Out}, pages 74--81, Barcelona, Spain. Association for Computational Linguistics.

\bibitem[{Lin et~al.(2020)Lin, Ji, Huang, and Wu}]{lin-etal-2020-joint}
Ying Lin, Heng Ji, Fei Huang, and Lingfei Wu. 2020.
\newblock \href {https://doi.org/10.18653/v1/2020.acl-main.713} {A joint neural model for information extraction with global features}.
\newblock In \emph{Proceedings of the 58th Annual Meeting of the Association for Computational Linguistics}, pages 7999--8009, Online. Association for Computational Linguistics.

\bibitem[{Liu et~al.(2023)Liu, Fabbri, Liu, Zhao, Nan, Han, Han, Joty, Wu, Xiong, and Radev}]{liu-etal-2023-revisiting}
Yixin Liu, Alex Fabbri, Pengfei Liu, Yilun Zhao, Linyong Nan, Ruilin Han, Simeng Han, Shafiq Joty, Chien-Sheng Wu, Caiming Xiong, and Dragomir Radev. 2023.
\newblock \href {https://doi.org/10.18653/v1/2023.acl-long.228} {Revisiting the gold standard: Grounding summarization evaluation with robust human evaluation}.
\newblock In \emph{Proceedings of the 61st Annual Meeting of the Association for Computational Linguistics (Volume 1: Long Papers)}, pages 4140--4170, Toronto, Canada. Association for Computational Linguistics.

\bibitem[{Min et~al.(2023)Min, Krishna, Lyu, Lewis, Yih, Koh, Iyyer, Zettlemoyer, and Hajishirzi}]{min-etal-2023-factscore}
Sewon Min, Kalpesh Krishna, Xinxi Lyu, Mike Lewis, Wen-tau Yih, Pang Koh, Mohit Iyyer, Luke Zettlemoyer, and Hannaneh Hajishirzi. 2023.
\newblock \href {https://doi.org/10.18653/v1/2023.emnlp-main.741} {{FA}ct{S}core: Fine-grained atomic evaluation of factual precision in long form text generation}.
\newblock In \emph{Proceedings of the 2023 Conference on Empirical Methods in Natural Language Processing}, pages 12076--12100, Singapore. Association for Computational Linguistics.

\bibitem[{Nguyen et~al.(2021)Nguyen, Lai, and Nguyen}]{nguyen-etal-2021-cross}
Minh~Van Nguyen, Viet~Dac Lai, and Thien~Huu Nguyen. 2021.
\newblock \href {https://doi.org/10.18653/v1/2021.naacl-main.3} {Cross-task instance representation interactions and label dependencies for joint information extraction with graph convolutional networks}.
\newblock In \emph{Proceedings of the 2021 Conference of the North American Chapter of the Association for Computational Linguistics: Human Language Technologies}, pages 27--38, Online. Association for Computational Linguistics.

\bibitem[{Touileb et~al.(2025)Touileb, Mikhailov, Kroka, {\O}vrelid, and Velldal}]{touileb2025norsumm}
Samia Touileb, Vladislav Mikhailov, Marie Kroka, Lilja {\O}vrelid, and Erik Velldal. 2025.
\newblock Benchmarking abstractive summarisation: A dataset of human-authored summaries of norwegian news articles.
\newblock In \emph{Proceedings of the Joint 25th Nordic Conference on Computational Linguistics and 11th Baltic Conference on Human Language Technologies(NoDaLiDa/Baltic-HLT 2025)}, pages 729--738, Tallinn, Estonia.

\bibitem[{Touileb et~al.(2024)Touileb, Murstad, M{\ae}hlum, Steskal, Storset, You, and {\O}vrelid}]{touileb-etal-2024-eden}
Samia Touileb, Jeanett Murstad, Petter M{\ae}hlum, Lubos Steskal, Lilja~Charlotte Storset, Huiling You, and Lilja {\O}vrelid. 2024.
\newblock \href {https://aclanthology.org/2024.lrec-main.488/} {{EDEN}: A dataset for event detection in {N}orwegian news}.
\newblock In \emph{Proceedings of the 2024 Joint International Conference on Computational Linguistics, Language Resources and Evaluation (LREC-COLING 2024)}, pages 5495--5506, Torino, Italia. ELRA and ICCL.

\bibitem[{You et~al.(2025)You, Touileb, Velldal, and {\O}vrelid}]{noreventgen}
Huiling You, Samia Touileb, Erik Velldal, and Lilja {\O}vrelid. 2025.
\newblock Noreventgen: generative event extraction from norwegian news.
\newblock In \emph{Proceedings of the Joint 25th Nordic Conference on Computational Linguistics and 11th Baltic Conference on Human Language Technologies(NoDaLiDa/Baltic-HLT 2025)}, pages 801--811, Tallinn, Estonia.

\bibitem[{Zhang and Bansal(2021)}]{zhang-bansal-2021-finding}
Shiyue Zhang and Mohit Bansal. 2021.
\newblock \href {https://doi.org/10.18653/v1/2021.emnlp-main.531} {Finding a balanced degree of automation for summary evaluation}.
\newblock In \emph{Proceedings of the 2021 Conference on Empirical Methods in Natural Language Processing}, pages 6617--6632, Online and Punta Cana, Dominican Republic. Association for Computational Linguistics.

\bibitem[{Zhang* et~al.(2020)Zhang*, Kishore*, Wu*, Weinberger, and Artzi}]{Zhang*2020BERTScore:}
Tianyi Zhang*, Varsha Kishore*, Felix Wu*, Kilian~Q. Weinberger, and Yoav Artzi. 2020.
\newblock \href {https://openreview.net/forum?id=SkeHuCVFDr} {Bertscore: Evaluating text generation with bert}.
\newblock In \emph{International Conference on Learning Representations}.

\end{thebibliography}

\appendix

\section{Summary statistics}

\begin{table*}[t!]
    \centering
    \begin{tabular}{l|ccc}
    \toprule
    \textbf{Summary} & \textbf{\#Summ.} & \textbf{\#Tokens} & \textbf{\#Avg.} \\ \midrule
    Annotator$_1$ & 33 & 8,679 & 263 \\
    Annotator$_2$ & 33 & 4,256 & 129 \\
    Annotator$_3$ & 33 & 2,732 & 83 \\ 
    \midrule
    Llama-3-8B-instruct & 33 & 3,308 & 100 \\
    Llama-3-8B & 33 & 4,331 & 131 \\
    Meta-Llama-3-8B-Instruct & 33 & 3,523 & 106 \\
    Mistral-Nemo-Instruct-2407 & 33 & 3,019 & 91 \\
    Normistral-11b-warm & 33 & 6,030 & 182 \\
    Normistral-7b-warm-instruct & 33 & 5,653 & 171 \\
    \bottomrule
    \end{tabular}
    \caption{Statistics of human-authored summaries and generated summaries for the test set of NorSumm. ``\#Summ.'': number of summaries; ``\#Tokens'': total number of tokens; ``\#Avg.'': average number of tokens per summary.}
    \label{tab:summ_stats}
\end{table*}

Writing summaries of news articles is a subjective task. Human annotators can write different summaries for the same article. In NorSumm, each article is accompanied with three unique summaries written different annotators, who write in very different styles. As shown in Table~\ref{tab:summ_stats}, Annotator$_1$ creates the longest summaries, while Annotator$_3$ creates the shortest summaries. The LLMs also generate varied summaries. As shown in Table~\ref{tab:summ_stats}, some models generate rather short summaries, while some models generate rather long summaries.




\end{document}